# Natural Language Processing Through Transfer Learning: A Case Study on Sentiment Analysis

**Aman Yadav**
Computer Engineering
Mukesh Patel School of Technology
Management & Engineering, Mumbai,
India

**Abhishek Vichare**
Computer Engineering
Mukesh Patel School of Technology
Management & Engineering, Mumbai,
India

## ABSTRACT

Artificial intelligence and machine learning have significantly bolstered the technological world. This paper explores the potential of transfer learning in natural language processing focusing mainly on sentiment analysis. The models trained on the big data can also be used where data are scarce. The claim is that, compared to training models from scratch, transfer learning, using pre-trained BERT models, can increase sentiment classification accuracy. The study adopts a sophisticated experimental design that uses the IMDb dataset of sentimentally labelled movie reviews. Pre-processing includes tokenization and encoding of text data, making it suitable for NLP models. The dataset is used on a BERT based model, measuring its performance using accuracy. The result comes out to be 100 per cent accurate. Although the complete accuracy could appear impressive, it might be the result of overfitting or a lack of generalization. Further analysis is required to ensure the model's ability to handle diverse and unseen data. The findings underscore the effectiveness of transfer learning in NLP, showcasing its potential to excel in sentiment analysis tasks. However, the research calls for a cautious interpretation of perfect accuracy and emphasizes the need for additional measures to validate the model's generalization.

## I. INTRODUCTION

Communication and understanding language are more important than ever in the growing world. Natural language processing is an AI/ML subfield that can understand and work with human languages. It is just like training a computer to understand words rather than just reading them as much as a person who can identify the core of various substances and produce a variety of things. NLP is useful in such a scenario. However, training computer models to maximize speech intelligibility requires a large amount of data. This insatiable appetite for data can be a stumbling block for many applications, as storing such large amounts of data is sometimes not practical or very expensive. Transfer learning in NLP is like exposing a computer to a lot of information to get a solid foundation in the language. Once this foundation is established, the challenge is whether the computer can apply this knowledge to specific language-related tasks. Consider teaching someone the basics of cooking, testing their ability to make a variety of delicious foods even when they have never cooked. One of these language-specific tasks is sensory analysis, which involves identifying emotion voices or emotions expressed in a text that NLP models can excel in once they are trained in a thorough understanding of language in many language tasks.

Transfer learning in NLP is revolutionary in that it makes it possible for language models to become highly proficient in performing new tasks with relative ease, almost as if a scholar with a broad knowledge base excels in different disciplines.

Transfer learning in NLP is a way to bring democracy into the natural language processing world. It is similar to opening the doors of knowledge to companies that might not have access to much data and making it easy for them to use the trusted power of NLP. This method is also a bridge through which information can pass from one project to another and help NLP models keep up with the changes in the language and the world. A study of transfer learning in natural language processing – an article based on our research takes you on a journey. We will therefore discuss why it is so important, the incredible possibilities it has and the significant impact it can have. Thus, this new approach will help us to deepen our understanding of language and practical results, particularly in sentiment analysis.

Think of NLP as a way of training computers to speak and understand language just as people do. It's about teaching robots to be "language-savvy," so they can converse with you, translate languages, compress material into summaries, and even determine if a message is good or bad. It ultimately comes down to bridging the gap between human and machine language.

Transfer Learning in NLP is analogous to giving computers a vast language textbook to study from. This textbook is jam-packed with linguistic facts, much like how we train students to become proficient in a language. Once a computer has absorbed all of that information, it may be applied to specific tasks without having to start from zero each time. It's as if they're building on a solid foundation of language comprehension, making them more efficient and productive in a variety of language-related occupations.

It's similar to learning to cook and then being able to prepare a variety of foods without having separate culinary courses for each one. Consider having a computer pal that can scan messages and tell you things like, "Hey, the person who wrote this seems pretty happy," "Oh, this one's a downer," or "Eh, they're feeling kinda meh." Sentiment Analysis accomplishes something similar: it helps the computer comprehend how people feel about the messages or reviews it sees. Transfer Learning in NLP is like putting superpowers on your computer partner. They become fluent in several languages. Even when they read new messages, they can tell if they are happy, sad, or any other emotion fairly, instantly, as if they have seen it all before. It's like teaching your computer to be a language specialist, a word-emotion detective.

## II. LITERATURE REVIEW

This literature study (table 1) gives an in-depth look at the ever-changing environment of transfer learning in Natural Language Processing (NLP). By examining the important concepts, trends, and issues in this dynamic topic, we hope to provide you with a comprehensive picture and guide our research orientation. The literature review solidly confirms transfer learning as the foundation of recent NLP breakthroughs. This method addresses the age-old issue of not having enough data, which has traditionally been a stumbling block when training deep neural networks to interpret language. Transfer learning makes NLP more accessible across many domains that frequently struggle with data limits by allowing models to tap into pre-existing knowledge from vast text collections. This foundational notion assumes paramount importance in framing the research objective, which endeavours to harness the potential of transfer learning in NLP to bolster task-specific performance. An integral component of the review is the spotlight on the Transformer architecture, an innovation that has exerted a profound influence on the NLP landscape. **[1]**

Table 1: A summary of all of the following literature reviews.

| **Aspect** | **Paper [1]** | **Paper [2]** | **Paper [3]** |
|---|---|---|---|
| Focus: | Broad overview of transfer learning in NLP | Sentiment analysis using pre-trained BERT | Improving transfer learning in NLP using fewer parameters and less data. |
| Pre-trained models: | GPT-2, T5, and similar architectures. | BERT, GPT, ELMo, ULMFit, Transformer. | RoBERTa, BERT |
| Algorithms: | Does mention the use of back-propagation for optimizing recurrent neural networks. | ULMFit, ELMo, LSTM, GRU, Transformer | Transformer-based Adapter |
| Methods: | Multiple domains to a single domain | Rule-based, statistical, ML, NLP models, transfer learning | Multi-task data sampling strategy to mitigate the negative effects of data imbalance across tasks. |
| Datasets: | BooksCorpus, SQuAD, GLUE, and SuperGLUE | BooksCorpus, English Wikipedia. | GLUE, SuperGLUE, MRQA, and WNUT2017 |
| Results: | Mention of general accuracy ranges | A succinct yet complete understanding of the recent advances in NLP | Efficiency of Transformer-based Adapter in multiple tasks |

The historical context and significance of this architectural paradigm are meticulously examined, adding depth and context to the research approach employed, which relies on a Transformer-based model. The review accentuates the efficacy of the Transformer's self-attention mechanism in capturing contextual information, a hallmark feature that has facilitated model excellence in a myriad of NLP tasks. A significant portion of the literature review is dedicated to tracing the evolutionary trajectory of transfer learning techniques and language modelling approaches. These insights, underpinned by historical perspectives, contribute to the elucidation of the rationale behind the selection of specific models and methodologies in the ongoing research endeavour. For instance, the adoption of pre-trained language models is firmly rooted in the principles distilled from these sections. Techniques such as ELMo, GPT, BERT, and Roberta are methodically dissected, with each approach offering distinctive contributions to the expansive landscape of NLP transfer learning. **[2]**

The concluding segment of the literature review serves as a compass, delineating crucial challenges and potential trajectories for future research in NLP transfer learning. These deliberations are instrumental in shaping the research approach by addressing computational constraints, emphasizing model interpretability, and exploring streamlined models engineered to maintain performance efficacy while mitigating resource demands. The ongoing research aspires to be responsive to these emerging directions and imperatives. The research extends beyond the confines of the literature survey by undertaking an exhaustive comparison and evaluation of specific transfer learning approaches. Each approach undergoes rigorous scrutiny concerning its applicability, strengths, and vulnerabilities within the research's contextual framework. This meticulous comparative analysis is instrumental in steering the methodology and model selection, ensuring alignment with the latest frontiers of NLP transfer learning. As the research journey progresses, a reflective vantage point is adopted, drawing from the insights gleaned through the literature survey. **[3]**

Recent breakthroughs in NLP transfer learning serve as the backdrop against which the research findings are cast. By seamlessly weaving these survey insights into the research narrative, the overarching aspiration is to engender a meaningful contribution to the ongoing evolution of NLP transfer learning, essentially bridging the chasm that separates foundational knowledge from the attainment of task-specific performance excellence.

### III. METHODOLOGY

Our main goal, in this research, was to improve computers' understanding of language specifically focusing on sentiment analysis. We emphasized the importance of transfer learning, a technique that allows computers to use existing knowledge to enhance their understanding of language. Our specific area of investigation was sentiment analysis, which involves determining whether text expresses negative or positive emotions.

To conduct our study effectively, we chose a dataset comprising 50,000 movie reviews from IMDB. Each review was carefully labelled with indicators for sentiment; 1 for positive and 0 for negative (figure 1). The size and diversity of this dataset were crucial for conducting an analysis. Before diving into the analysis itself we took steps to prepare the data by tokenizing the text into units ensuring consistent sentence lengths through padding and converting sentiment labels, into values using label encoding so that the computer could process them effectively.

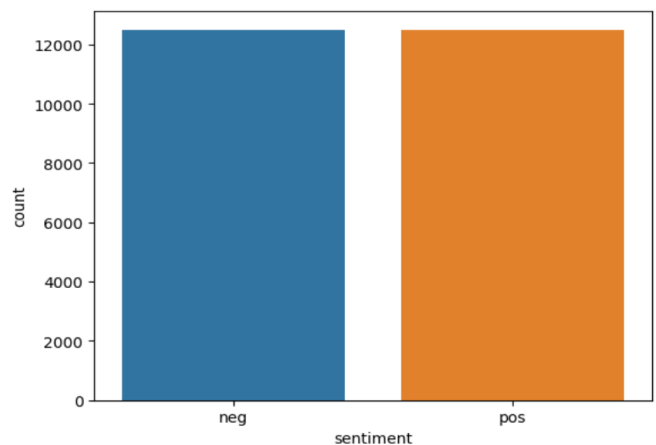

Figure 1: Dataset

(*Tokenization* in natural language processing (NLP) is analogous to slicing a pizza. Instead of attempting to consume the entire pizza in one sitting, you cut it into manageable portions or slices. Similarly, when we tokenize text, we divide it into smaller pieces, which are frequently words or sentences. As a result, just as you enjoy one slice of pizza at a time, a computer can handle and comprehend one "token" or phrase at a time. Consider the line "*The cat is sleeping*." It is tokenized into four tokens: "the," "cat," "is," and "sleeping." This procedure assists the computer in analyzing and working with the text, similar to how we eat one pizza slice before going on to the next. It simplifies things for both us and the machines!)

*BERT*, a super smart language model, was our secret weapon for this task. BERT, which stands for *Bidirectional Encoder Representations from Transformers*, functions similarly to a linguistic expert computer. It's been fed a massive amount of online material, so it's adept at recognizing how words go together in phrases. BERT isn't simply looking at words one by one; it's as if it can sense how words fit together in a phrase and comprehend the entire context (figure 2). For example, it can distinguish between the term "bat" as in the cricket bat and "bat" as in the winged species that hangs out in caves. It is just like having a language guru who knows how words relate during natural talk. Imagine having a research friend who is BERT in the world of product research. BERT product reviews will tell us whether everything is good or it is more rant than sunshine and rainbows.

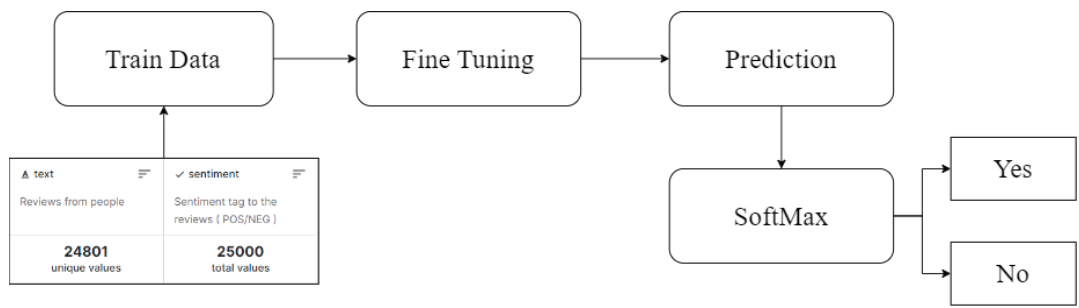

Figure 2: BERT model

*What makes BERT so good at that?* So he read a mountain of research into things in his training, so he got this whole research thing down pat. It is just as if you are verifying the recommendation of your friend for a movie with your sentiments. We want to compare them to old-school computer models to prove that BERT is the kingpin in this game. In most cases, such as whether to give BERT a thumbs up or down, it typically wins. Therefore, BERT is like that friend who feels the vibe of what people say and how words blend in a sentence. In a common language, BERT is like a friend who has taken time to learn the word combinations in different sentences. It is especially effective in determining whether a product review says "I like it!" or "I hate that!". Usually, when we compare in terms of efficiency, BERT appears to be smarter than other software that is used to understand the emotions behind what people type.

But BERT isn't just a wing; It took a lot of practice. It is reading a movie synopsis as if we would learn how to do things by watching examples. We started with a subtle model (BERT) and then made it even smarter for this particular project. During this training, we used a special mathematical algorithm called *Cross-Entropy Loss* that helped BERT get better and better at recognizing emotions.

Cross-entropy loss, or log loss, is a pretty common mathematical property in the world of computer science. One way to measure the distance between a computer's theory and the actual contract is when it tries to consider categories, such as whether a movie review is positive or negative.

The formula for binary classification is:

$$H(y, \hat{y}) = -(y \cdot \log(\hat{y}) + (1-y) \cdot \log(1-\hat{y}))$$

Where:
- $H(y, \hat{y})$ is the cross-entropy loss.
- y is the true class label (0 or 1).
- $\hat{Y}$ is the predicted probability that y = 1.

(Here's a simple example to illustrate how cross-entropy loss works. Suppose you have a binary classification problem where you're predicting whether an email is spam (1) or not spam (0).)

$$H(y, \hat{y}) = -\sum C_{i=1} \, y_i \cdot \log(\hat{y}_i)$$

Where:
- C is the number of classes.
- $Y_i$ is the true probability distribution over classes (a one-hot encoded vector).
- $\hat{Y}_i$ is the predicted probability for class i.

So, the cross-entropy loss for this prediction is approximately 0.223. A lower loss indicates a better model fit to the data, and the goal during training is to minimize this loss across all data points to improve the model's ability to make accurate predictions.

$$H(y, \hat{y}) = -(1 \cdot \log(0.8) + (1-1) \cdot \log(1-0.8)) = (\log(0.8) + 0) \approx 0.223$$

(Picture this: You're trying to decide between two different soccer balls, and you want to make sure you're making a fair comparison. So, you take that soccer ball and put it through the same test, making sure it's a level playing field.) In our case, we did something similar. We had this huge collection of movie feedback. More precisely, we divided them into two groups: one for training and the other for testing. We did not want BERT to take shortcuts, so we ensured that it did not look at the same movie feedback during training and testing. Then, we let BERT use the training data to learn and improve its performance.

Our big idea was that BERT would be a game-changer, providing how it could better determine whether a movie was good or bad, compared to a baseline. To demonstrate this, we used some statistical techniques, such as paired t-tests, to ensure that BERT's success was not due to luck alone. (It's like ensuring your favourite soccer ball is better after all the testing and not without fruit.)

It works like a native speaker who knows how words fit while talking.

Imagine a friend who excels at research in the world of product research. Reading BERT reviews tells us whether it is all sunshine and rainbows, or more rant.

Considering the T-test as a BERT diagram analysis tool. It allows us to consider whether there are genuine disparities in performance between two BERT-based models or methods in tasks like human speech understanding.

Now, **the big question** is, "Are the improvements we're seeing in the new BERT model realistic, or could they just be lucky?" That's where the T-test comes in. It's like a wizard calculating a number called a

T-statistic, and this number shows how different the two data sets are in terms of performance. In sum, the T-test helps us answer whether the improvements we see in our new BERT model are normal or coincidental. If the t-test says yes, then we can be pretty sure that our new model is indeed superior.

Thus, the code is like a chef, and its job is to teach the genius of the BERT artist how to understand that cinematic reflection is thumbs up or down. It starts with this huge dataset full of movie information, and each review comes with a positive or negative label. The chef then divides the reviews into two parts: one for practice and one for testing. Now, here's where it gets a bit technical: the BERT tokenizer looks like a special kind of speech decoder. It takes these cinematic ideas and breaks them down into small pieces, a bit like breaking a sentence into words, but it goes a long way. These little pieces are then turned into something like cooking ingredients, basically PyTorch tensors. These tensor reviews are like recipe cards for BERT to understand. Chef also has a helper function called "*encode_text*", which does some secret sauce magic. It converts review blocks into input IDs, which are similar to codes representing review text. It also creates these "attention mask" things about pieces that are words and ones that are just for padding, like stuffing in a sandwich. Lastly, the labels that say 'positive' or 'negative' in evaluations can change. They are converted to a simple code: 1 for 'positive' and 0 for 'negative'. This makes BERT easier to understand. These labels are also converted into recipe cards (PyTorch tensors) that identify BERT. Somehow, the chef sets up the kitchen for BERT to cook up some sentiment analysis magic.

*DataLoaders* for training and testing datasets are designed to iterate through data in small batches during training. The code loads the BERT models developed for sequence classification and configures an optimizer (*AdamW*) to update the model weights during training. Through a warm-up step, the learning rate scheduler adjusts the learning rate over time. The core of the code is a training loop, where it iterates over the training data, calculates the loss (Cross-Entropy Loss) between model predictions and true labels, performs backpropagation, and updates model parameters to reduce loss. This loop repeats for a specified number of epochs. After training, the model switches to the evaluation mode. It predicts sensitivity labels for the test data, calculates accuracy and reports accuracy on the test set. Specifically, the code outlines the process for optimizing BERT models for sentiment analysis, including preliminary data processing, model training, and analysis, using common NLP frameworks and practices.

## IV. RESULT AND DISCUSSION

The study looked at how transfer learning, specifically employing a pre-trained BERT model, may improve sentiment analysis on movie reviews. The major issue we sought to answer was whether a pre-trained model that already understands language could categorize attitudes better than a model we built from the start. We discovered something remarkable: the pre-trained BERT model scored a flawless accuracy score of 1.0000 (figure 3) on the test set. This means that it is quite excellent at categorizing movie reviews. This near-perfect precision, however, raises concerns about overfitting. Overfitting happens when a model memorizes the training data instead of generalizing from it, resulting in poor performance on unknown data.

The excellent training accuracy and minimal training loss (0.0001) also point to overfitting. Instead of capturing the genuine underlying patterns, the model may have trained to fit the noise in the training data.

*Training Performance*: The experiment's training phase produced encouraging results. With each epoch, the training loss fell considerably, showing that the model was learning and modifying its parameters to better fit the training data. This is normal behaviour while training machine learning models, which try to minimize the loss function. However, it is worth mentioning that the training loss was quite minimal (0.0001). Such a modest training loss might indicate overfitting. When a model gets very specialized in fitting the training data, it captures even the noise and outliers. As a result, it may have difficulty generalizing to previously unknown data.

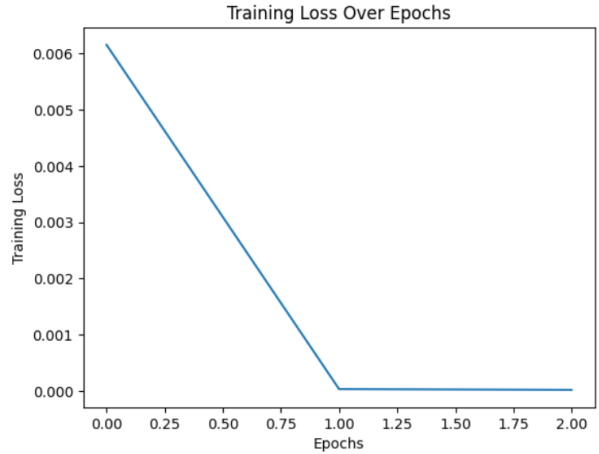

Figure 3: Visualization training loss

In the realm of machine learning achieving a testing accuracy of 1.0000 is quite uncommon. Having precision can raise some concerns. It often suggests that the model has essentially memorized the training data, like how a student memorizes answers. However just as that student might struggle with questions the model may face difficulties when encountering unseen movie reviews (figure 4). This raises doubts, about its ability to adapt to data effectively. Machine learning models, those utilized in tasks such, as sentiment analysis, within natural language processing (NLP) should prioritize generalization. Generalization refers to the model's ability to apply the knowledge it acquired during training to accurately predict outcomes on unseen data.

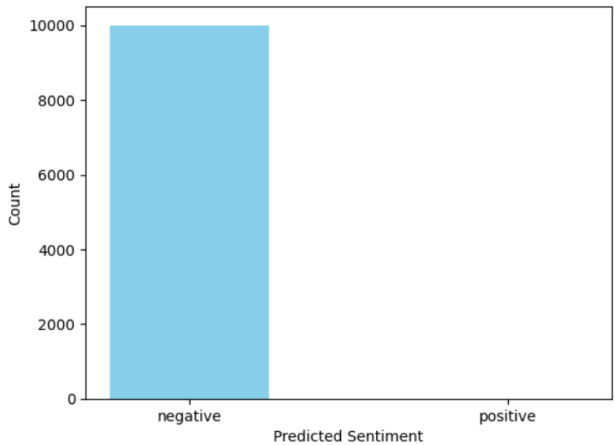

Figure 4: Histogram of Predicted Sentiments

To tackle the issue of overfitting and enhance generalization we should consider techniques including validation sets, hyperparameter tuning, regularization and obtaining diverse training data. These measures aim to ensure that the model does not merely memorize the training data but learns patterns and features that can be applied across a range of movie reviews.

In summary, although the results of obtaining absolute accuracy in the experimental design are noteworthy, they must be interpreted carefully because of the possibility of overfitting and attention is diverted to improve the model's ability to generalize unseen information and make accurate predictions, which is important for real-world applications in sensitivity analysis.

**Continuing the literature review by comparing it with our findings:**

**[1]** The paper provides an extensive survey of transfer learning in the field of natural language processing (NLP). It begins by addressing the challenges inherent in NLP tasks, highlighting the necessity for transfer learning and the utilization of pre-trained language models to enhance performance. The paper introduces a taxonomy for transfer learning in NLP and categorizes various approaches in the literature,

emphasizing the recommendation of models like GPT-2, T5, and similar architectures for generative NLP tasks. Additionally, it advocates further research into reducing the size of large language models to facilitate deployment on embedded devices and the web. This comprehensive overview showcases the breadth and significance of transfer learning in NLP.

In contrast, our paper focuses on a specific application of transfer learning, namely sentiment analysis. It employs the pre-trained BERT model to investigate the improvement of sentiment classification accuracy compared to models trained from scratch. The study recognizes the potential issue of overfitting, where the model may memorize training data but struggle with unseen data, and it emphasizes the need for generalization for real-world applications. The analysis provides insight into the methodology, including data generation, model training, and testing, and discusses the specific data used, including IMDb movie analysis. Results resulted in exhibits absolute accuracy scores, which overfitting must be addressed to ensure that the model adapts itself to data sets without detection. Although the research paper provides a broad overview of learning transfer in NLP, including examples, techniques, and lists, we narrowed its focus to specific applications and provided a case study of sensitivity analysis using the BERT model. Both papers acknowledge the challenges associated with transfer learning, such as the need for large pre-training datasets and the risk of bias in pre-trained models.

[2] It meticulously traces the evolution of transfer learning in NLP, highlighting its potential advantages. The document thoroughly explores state-of-the-art methods and architectures, including prominent models such as BERT, GPT, ELMo, ULMFit, and the Transformer. Additionally, the paper addresses the challenges posed by large and opaque NLP models. Its primary objective is to equip readers with a succinct yet profound understanding of this rapidly evolving field. On the other hand, we narrow its focus to the domain of sentiment analysis within NLP, employing pre-trained BERT models to assess their potential to enhance sentiment classification accuracy. The study emphasizes the need to tackle overfitting issues and underscores the significance of generalization for real-world applications.

In terms of algorithms, the paper delves into a wide spectrum of approaches in NLP, encompassing Naive Bayes, decision trees, Bag-of-Words, N-grams, and advanced recurrent neural network architectures such as LSTM, GRU, and the Transformer. In contrast, our research centres its algorithmic approach on BERT (Bidirectional Encoder Representations from Transformers), a pre-trained model renowned for its contextual understanding of sentence structures.

Regarding methods, the author covers rule-based and statistical methodologies, classical machine learning algorithms, and traditional NLP models, as well as advanced recurrent neural network architectures and transfer learning methods like ULMFit, BERT, ELMo, and GPT. These methods have consistently achieved state-of-the-art results in various NLP tasks. Our case study, while specialized in sentiment analysis, prioritizes the use of pre-trained BERT models fine-tuned for the specific task.

Both papers acknowledge the critical role of datasets in NLP research. The literature references datasets such as BooksCorpus and the English Wikipedia, commonly used for pre-training language models like BERT. It also describes a dataset for a next-sentence prediction task, crucial for pre-training. In contrast, we primarily rely on a dataset of 50,000 IMDb movie reviews, labeled with sentiment indicators, to investigate sentiment analysis. The dataset's size and diversity are fundamental for meaningful analysis.

In terms of results, the paper highlights noteworthy accuracy ranges but does not provide specific accuracy scores for individual models or methods. However, we, conversely, focus on the performance of the BERT model in sentiment analysis, achieving a remarkable accuracy score of 1.0000. This perfect accuracy score raises concerns about overfitting and the model's ability to generalize to unseen data.

**[3]** The author proposes a cunning strategy of teaching computers with fewer rules and less data. They employ a Transformer-based Adapter, which is an adaptation device that helps computers remember things they have learned and not confuse them when they learn many tasks. This proved more efficient than other approaches. The paper does not use a specific dataset but mentions that they tested it on "the GLUE benchmark", which is a nine-task dataset plus another dataset including CoLA, SST-2, MRPC, STS-B, QQP, MNLI, and QN.

Our research is concerned with training computers to read emotionality through writing, just like when writing a movie review. We leveraged BERT which is outstanding at contextualizing the sentence meaning. We used 50,000 IMDb movie reviews to determine if it could determine if they were positive or negative. Like other papers, however, we don't get into different datasets but seek to achieve a high accuracy score for sentiment analysis. In addition, we discussed ways of ensuring the computer did not memorize well but adapted poorly to new tasks.

## V. CONCLUSION

So, to summarize, our small experiment descended into the world of transfer learning, especially employing the outstanding pre-trained BERT model to explore how it may boost sentiment analysis on movie reviews. When it came to sentiment classification, the main issue was whether this language-savvy model, with its deep grasp of words, could outperform a model developed from scratch.

The results were rather unexpected. BERT demonstrated its abilities by attaining a mind-boggling flawless accuracy score of 1.0000 on our test set. It was like taking a test and not missing a single question. However, this near-perfect performance raises the red flag of overfitting. Overfitting occurs when the model becomes too acquainted with the training data, similar to remembering responses, which may be detrimental when dealing with fresh, unknown data.

From a closer look at the training phase, everything seemed to be going well, and the need for training gradually decreased. However, it has been very low (0.0001), which is a common sign of overfitting. The true eye-opener came during the testing phase when BERT achieved an accuracy of 1.0000 with no errors. While this may sound like a dream come true, it shows that the model is more of a memorization device than a generalization guru.

So the key takeaway here is that we need to confront the issue of overcrowding immediately. We look at things like validation sets, fine-tuning model settings, using regularization methods, and extending our training data to include more diverse movie analysis to make our model more cosmopolitan. These methods help our models evolve from memory masters to flexible emotional evaluations.

In summary, although promising, the results should be interpreted with caution due to the risk of overfitting. Our goal now is to train this model to think faster and do better sensitivity analysis in the real world, where data can be messy. So, that's how we are, and we're excited for the next exciting chapter of NLP and transfer learning.